\documentclass[acmtog]{acmart}
\acmSubmissionID{1414}

\usepackage{booktabs} %
\usepackage{multirow}

\usepackage[ruled]{algorithm2e} %

\SetAlFnt{\small}
\SetAlCapFnt{\small}
\SetAlCapNameFnt{\small}
\SetAlCapHSkip{0pt}

\acmJournal{TOG}

\begin{document}
\title{UVFaceFusion: Fast Multi-view Topologically Consistent Face Reconstruction in the Wild via UV-space Neural Fusion}

\author{Xin Ming}
\affiliation{%
  \institution{School of Software and BNRist, Tsinghua University}
  \city{Beijing}
  \country{China}
}
\email{1729406968@qq.com}
\orcid{0009-0007-9602-6078}

\author{Yuxuan Han}
\affiliation{%
  \institution{School of Software and BNRist, Tsinghua University}
  \city{Beijing}
  \country{China}
}
\email{hanyx22@mails.tsinghua.edu.cn}
\orcid{0000-0002-2844-5074}

\author{Junhai Yong}
\affiliation{%
  \institution{School of Software and BNRist, Tsinghua University}
  \city{Beijing}
  \country{China}
}
\email{Yongjh@tsinghua.edu.cn}
\orcid{0000-0002-4326-4167}

\author{Feng Xu}
\affiliation{%
  \institution{School of Software and BNRist, Tsinghua University}
  \city{Beijing}
  \country{China}
}
\email{xufeng2003@gmail.com}
\orcid{0000-0002-0953-1057}

\begin{abstract}
Reconstructing high-fidelity facial geometry with an assigned topology is essential for digital avatar creation and animation, yet existing automated methods often trade off geometric fidelity and in-the-wild generalization. 
We present UVFaceFusion, a feed-forward framework for multi-view, fixed-topology face reconstruction from daily images. 
Our key idea is to replace heuristic topological optimization with learnable neural fusion in a canonical UV space. 
Given multi-view images, we first obtain dense point maps and facial UV correspondences of each view using VGGT and Pixel3DMM, respectively. 
Then, the view-specific point maps are lifted into the canonical UV domain
and fused with a novel mask-aware neural fusion network. 
The network predicts a complete UV-space point map, from which a fixed-topology mesh is directly sampled.
Although trained only on Ava-256, UVFaceFusion generalizes well to multiple public benchmarks and in-the-wild captures, benefiting from its canonical UV-space geometry-to-geometry fusion that reduces dependence on dataset-specific appearance and capture conditions. 
Experiments on various benchmarks show that UVFaceFusion achieves state-of-the-art reconstruction accuracy while reconstructing a mesh from 16 input views in less than 3 seconds on a single RTX 4090.
Code is available at https://github.com/grignarder/UVFaceFusion.
\end{abstract}

\setcopyright{cc}
\setcctype{by}
\acmJournal{TOG}

\ccsdesc[500]{Computing methodologies~Reconstruction}

\begin{teaserfigure}
\centering
    \includegraphics[width=1.0\textwidth]{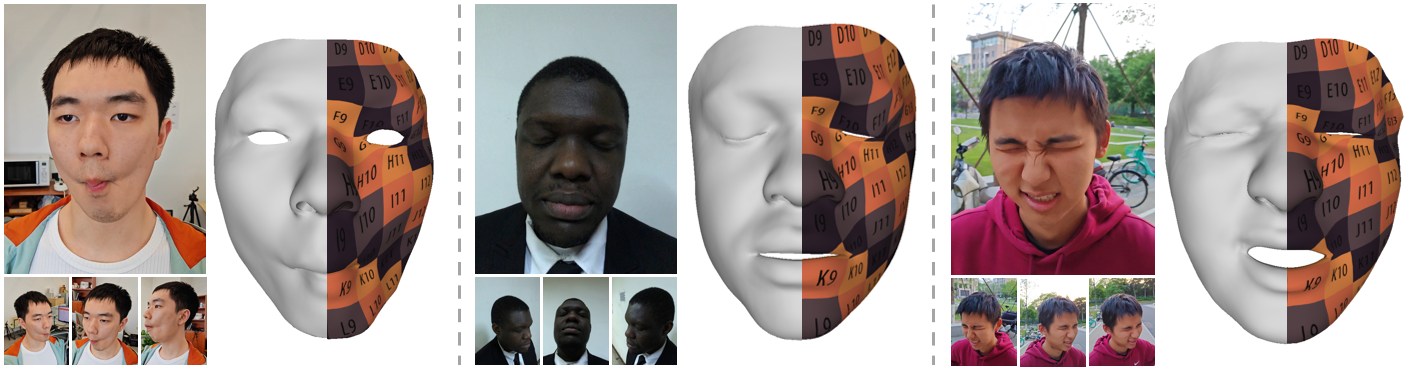}
    \caption{
    Given casual in-the-wild multi-view captures as input, with four of the sixteen images shown here, our method reconstructs a high-fidelity mesh with consistent topology in under three seconds. 
    It generalizes robustly across capture environments, lighting conditions, facial expressions, and ethnic groups, enabling everyday users to bring themselves into the digital world through casual scanning. 
    }
    \label{fig:teaser}
\end{teaserfigure}

\maketitle

\section{INTRODUCTION}

Establishing facial geometry with an assigned topology is a fundamental requirement for digital avatar creation~\cite{alexander2009digital,alexander2013digital}, as dense surface correspondences are required to facilitate rigging, animation, and texture transfer. 
Conventional paradigms generally separate the process into face geometry reconstruction followed by template fitting~\cite{egger20203d,riviere2020single}. 
However, ensuring high-fidelity output usually necessitates extensive manual labor for the second step, such as landmark supervision and meticulous parameter tuning, which restricts the scalability of these pipelines for widespread consumer applications. 
On the other hand, 3D Morphable Model (3DMM)-based methods~\cite{deng2019accurate,DECA:Siggraph2021} can be used for automatic template fitting but are inherently limited by the low-dimensional expression space, failing to capture person-specific geometric traits.  
To pursue higher fidelity, multi-view learning approaches~\cite{li2021tofu,li2024grape} employ neural networks to infer geometry directly from images. 
However, these methods essentially couple geometry reconstruction with topology building, making them heavily reliant on studio-captured datasets for end-to-end training. 
Consequently, they struggle to generalize to in-the-wild scenarios where lighting and camera configurations deviate from the controlled training environment.

To resolve this generalization-fidelity trade-off, VGGTFace~\cite{ming2026vggtface} proposes a \emph{decoupled} paradigm that decomposes topologically consistent reconstruction into two stages: general geometric reconstruction and topological fusion.
Given smartphone-captured multi-view face images in the wild, VGGTFace leverages a robust, general-purpose estimator (VGGT~\cite{wang2025vggt}) for geometry prediction (in terms of view-specific point maps), which achieves high generalization across diverse environments.
Then, it applies a heuristic Bundle Adjustment (BA)~\cite{triggs1999bundle} process to integrate these view-specific predictions into a unified, topologically consistent mesh.
However, this hand-crafted BA optimization lacks semantic awareness to distinguish noise in geometry predictions from actual facial features, often resulting in an over-smooth surface or topological artifacts that fall short of the precision required by professional pipelines.

To close this gap, we present UVFaceFusion, which evolves the decoupled paradigm by replacing the rigid, heuristic BA with a learnable neural fusion module.
To inherit the high generalization ability of general-purpose estimators, we operate exclusively on geometric point maps, effectively shielding the fusion process from the detrimental effects of lighting variations and camera distortion that are pre-processed by the underlying VGGT backbone. 
Specifically, we utilize the predicted UV correspondence by Pixel3DMM~\cite{giebenhain2025pixel3dmm} to warp view-specific 3D coordinates into a canonical UV space, thereby transforming the irregular multi-view fusion problem into a structured, pixel-aligned feature aggregation task. 
Then, we employ a shared-weight encoder to extract deep geometric feature maps from these aligned point maps, followed by a mask-aware feature fusion layer that dynamically aggregates cross-view information to resolve occlusions and prediction uncertainties. 
Finally, a decoder reconstructs the refined, topologically consistent facial geometry in the UV domain. 
By focusing on geometry-to-geometry mapping, our framework demonstrates exceptional data efficiency; it can be supervised on limited studio datasets, such as Ava-256~\cite{martinez2024ava256}, yet generalizes seamlessly to unconstrained in-the-wild faces. 
Notably, given 16 input views, UVFaceFusion reconstructs a fixed-topology face mesh in less than 3 seconds on a single RTX 4090, achieving a 3$\times$ inference speedup over VGGTFace while setting new benchmarks in both qualitative fidelity and quantitative accuracy.

In conclusion, our main contributions include:
\begin{itemize}
    \item We present an efficient feed-forward system that reconstructs face meshes with fixed topology from 16 in-the-wild images in less than 3 seconds on a single RTX 4090, with a 3$\times$ speedup over VGGTFace and state-of-the-art accuracy across multiple benchmarks.
    
    \item We introduce a canonical UV-space fusion representation that lifts VGGT point maps into a shared facial parameterization using Pixel3DMM correspondences, converting fixed-topology face reconstruction into a structured UV point-map prediction problem.

    \item We propose a mask-aware neural fusion network that first extracts per-view geometric features with a shared encoder and then aggregates them in feature space, enabling robust fusion of noisy and partial multi-view observations.
\end{itemize}

\section{RELATED WORK}
\subsection{Monocular Face Reconstruction}
Reconstructing 3D facial geometry from a single image is inherently ill-posed, and a large body of work addresses this ambiguity with 3D Morphable Model (3DMM) priors~\cite{DBLP:conf/siggraph/BlanzV99,FLAME:SiggraphAsia2017,zollhofer2018state,egger20203d}.
Given an input image, these methods either solve an analysis-by-synthesis fitting problem~\cite{thies2016face2face,dib2021practical,giebenhain2025pixel3dmm} or regress the parameters of a 3DMM/FLAME model with a neural network~\cite{deng2019accurate,DECA:Siggraph2021,zielonka2022mica,danecek2022emoca,wang20243d,retsinas2024smirk}.
Recent methods further improve robustness to challenging expressions and in-the-wild images by using denser supervision such as facial-part segmentation~\cite{wang20243d}, neural rendering objectives~\cite{retsinas2024smirk}, or foundation-model features for screen-space normal and UV prediction~\cite{giebenhain2025pixel3dmm}.
While these approaches produce meshes with a consistent template topology, their final geometry is still largely governed by the underlying parametric face space and the ambiguity of a single observation.

To improve geometric expressiveness, many works augment the parametric model with per-vertex offsets, UV displacement maps, or learned nonlinear corrective spaces~\cite{DECA:Siggraph2021,yang2020facescape,lei2023hierarchical,dib2024mosar,tewari2019fml,han2023ReflectanceMM}.
These extensions recover more local details, yet they typically treat high-frequency geometry as residuals on top of a coarse 3DMM fit.
Another line of work reduces the dependence on 3DMM parameters by directly predicting dense geometry or correspondences in image/UV space~\cite{richardson2017learning,sela2017unrestricted,feng2018joint,guo2023rafare,Zhang_2021_CVPR,yun2025warphe4d}.
For example, \citet{sela2017unrestricted} estimate a UV coordinate image and a depth map and then fit a template mesh to obtain topologically consistent geometry, while later methods predict UV position maps, implicit fields, or dense UV-depth maps.
These representations are closely related to ours because they expose dense correspondences and geometric maps rather than only low-dimensional coefficients.
However, monocular methods must hallucinate self-occluded regions from learned priors, and per-image predictions do not exploit multi-view observations that are available in our setting.

In contrast, our method treats monocular face reconstruction modules as providers of partial per-view cues instead of final reconstruction engines.
We use Pixel3DMM to obtain pixel-aligned UV maps and combine them with VGGT point maps~\cite{wang2025vggt}; after warping them into a shared UV domain, our networks learn to fuse masked partial observations from multiple views and predict a complete UV-space point map.
This design preserves the fixed topology required by downstream facial animation pipelines, while freeing the final geometry from the low-dimensional 3DMM space and reducing the hallucination burden inherent to single-image reconstruction.

\subsection{Multi-view Face Reconstruction}
Multi-view observations provide complementary geometric cues and largely reduce the ambiguity of monocular reconstruction.
Traditional production pipelines usually first reconstruct an unstructured facial scan using multi-view stereo or photogrammetry, and then register a template mesh to obtain dense correspondence~\cite{fyffe2017multi,riviere2020single}.
Although such pipelines can achieve high fidelity, they typically require controlled capture setups, accurate calibration, and substantial manual post-processing.
Some learning-assisted methods formulate multi-view face reconstruction as 3DMM regression or non-rigid multi-view stereo~\cite{wu2019mvf,bai2020deep}.
By enforcing cross-view consistency, these methods improve over monocular reconstruction, but their geometry is still constrained by parametric or adaptive face spaces and thus has limited ability to represent person-specific details.

Recent works directly infer topologically consistent meshes from calibrated multi-view images.
ToFu~\cite{li2021tofu} predicts template vertices with volumetric sampling, ReFA~\cite{liu2022refa} performs recurrent feature alignment in the UV space to recover production-grade face assets, and TEMPEH~\cite{bolkart2023tempeh} learns to predict dense-correspondence head meshes with view- and surface-aware volumetric feature fusion.
GRAPE~\cite{li2024grape} further improves the generalization of such systems to unseen camera arrays by using visual-hull-based localization and visibility-aware feature aggregation.
Very recently, MOCHI~\cite{filntisis2026mochi} removes the need for precomputed registered meshes during training by using differentiable point-map and normal-map losses with a topology-regularizing inverse-kinematics branch.
Despite their impressive accuracy, these methods are mainly designed for calibrated studio captures and often rely on carefully processed scans or registered meshes for supervision.
Their generalization to casual in-the-wild captures remains challenging.

Another relevant line of work predicts dense screen-space geometric maps that can be used for multi-view integration.
WarpHE4D~\cite{yun2025warphe4d} predicts a dense 4D head map, including per-pixel UV coordinates, depth, and confidence, enabling multi-view point-cloud fusion and downstream template registration.
This representation is closely related to ours because it exposes dense correspondences between image pixels and a shared UV domain.
However, WarpHE4D predicts each view independently and leaves the fusion of incomplete and noisy observations to downstream processing.
In contrast, our method learns to aggregate partial UV-space observations from multiple views into a complete fixed-topology facial geometry.

The most relevant work to ours is VGGTFace~\cite{ming2026vggtface}, which introduces a decoupled paradigm for in-the-wild multi-view facial geometry reconstruction.
It first leverages VGGT~\cite{wang2025vggt} to predict view-specific point maps and then injects topology with Pixel3DMM~\cite{giebenhain2025pixel3dmm}.
The resulting per-view topological point clouds are fused by a topology-aware Bundle Adjustment process.
While this design inherits the strong generalization ability of VGGT, its fusion stage is a hand-crafted optimization based on reprojection and Laplacian regularization.
Such an optimization lacks semantic awareness to distinguish noisy tracks from genuine facial structures, and may therefore over-smooth high-frequency details or produce local artifacts.
Our method inherits the robustness of foundation-model point maps for in-the-wild geometry estimation, while replacing heuristic test-time fusion with a learned reconstruction framework in canonical UV space. 
Specifically, we lift view-specific point maps into a shared facial UV parameterization and aggregate the resulting masked partial observations through neural fusion. 
This geometry-to-geometry formulation enables efficient fixed-topology reconstruction and better preserves person-specific facial shape, expressions, and local details.

\section{METHOD}
\begin{figure*}[t]
    \centering
    \includegraphics[width=1.\textwidth]{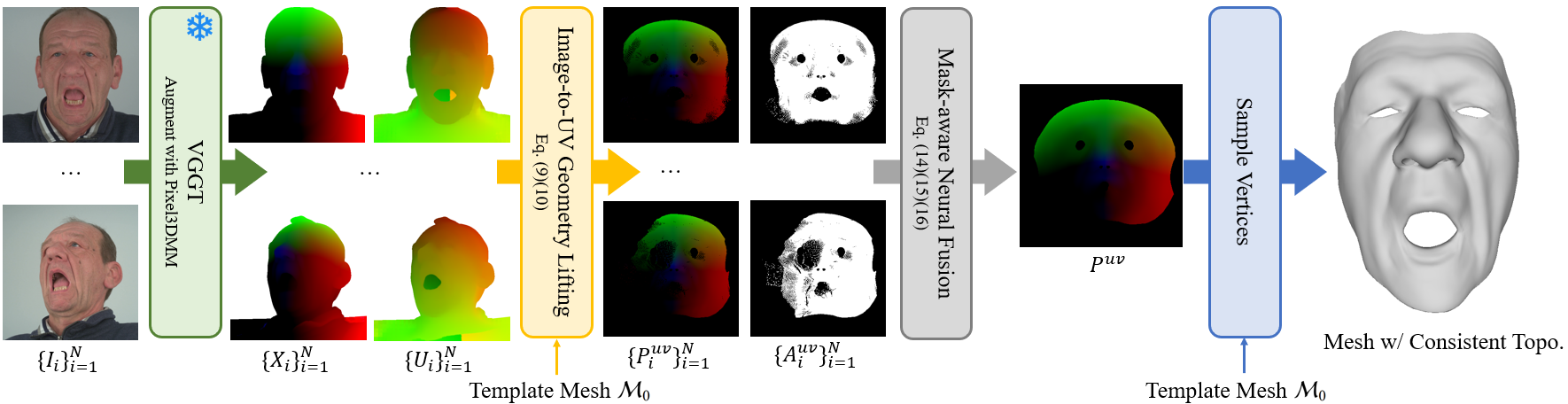}
    \caption{
    \textbf{Pipeline overview.}
    Given multi-view face images $\{I_i\}_{i=1}^{N}$, we use frozen pretrained models to predict pixel-aligned point maps $\{X_i\}_{i=1}^{N}$ and dense UV correspondences $\{U_i\}_{i=1}^{N}$.
    Through image-to-UV geometry lifting and canonicalization with the template mesh $\mathcal{M}_0$, each view is converted into a partial canonical UV-space point map $\{P_i^{uv}\}_{i=1}^{N}$ with a valid mask $\{A_i^{uv}\}_{i=1}^{N}$.
    Our mask-aware neural fusion network aggregates these partial observations and predicts a complete UV-space point map $P^{uv}$.
    Finally, we sample mesh vertices from $P^{uv}$ using the template UV coordinates and reuse the template connectivity to obtain a mesh with consistent topology.
    The snowflake indicates frozen pretrained modules.
    }
    \label{Fig:framework}
    \vspace{-0.5em}
\end{figure*}

As shown in Figure~\ref{Fig:framework}, given multi-view face images, we first use VGGT to estimate pixel-aligned point maps and Pixel3DMM to predict dense UV correspondences. 
We lift each view's point map into the canonical facial UV domain, yielding partial UV-space geometry observations.
A mask-aware neural fusion network then aggregates these partial observations in feature space and predicts a complete UV-space point map. 
Finally, the fixed-topology mesh is sampled from the UV-space point map.

\subsection{Preliminaries}
\label{sec:preliminaries}

Given a set of multi-view face images 
$\mathcal{I}=\{I_i\}_{i=1}^{N}$, our goal is to reconstruct a 3D face mesh with a fixed topology. 
Instead of directly regressing mesh vertices, we represent the output geometry as a dense point map in a canonical facial UV space. 
The final mesh is obtained by sampling this UV-space point map according to the UV parameterization of a template mesh.

\paragraph{Template mesh and UV parameterization.}
Let
\begin{equation}
    \mathcal{M}_0 = (\bar{\mathbf{V}}, \mathbf{F}, \bar{\mathbf{U}})
\end{equation}
denote the canonical template mesh, where 
$\bar{\mathbf{V}}\in\mathbb{R}^{M\times 3}$ are the template vertices, 
$\mathbf{F}$ are the fixed triangle faces, and 
$\bar{\mathbf{U}}=\{\bar{\mathbf{u}}_j\}_{j=1}^{M}$ are the corresponding vertex UV coordinates with $\bar{\mathbf{u}}_j\in[0,1]^2$.
The topology $\mathbf{F}$ is shared by all reconstructed meshes. 
Our network predicts a UV-space point map
\begin{equation}
    \mathbf{P}^{uv}\in\mathbb{R}^{S\times S\times 3},
\end{equation}
where each valid UV texel stores a 3D point on the facial surface in the canonical coordinate system.
Given $\mathbf{P}^{uv}$, we obtain the fixed-topology mesh by sampling the point map at the template UV coordinates $\bar{\mathbf{U}}$.

\paragraph{VGGT predictions.}
We use VGGT~\cite{wang2025vggt} to obtain dense multi-view geometric observations and camera parameters. 
Given the input images $\{I_i\}_{i=1}^{N}$, VGGT predicts
\begin{equation}
    \{ \mathbf{X}_i, \mathbf{K}_i, \mathbf{E}_i \}_{i=1}^{N}
    =
    \Phi_{\mathrm{VGGT}}(\{I_i\}_{i=1}^{N}),
\end{equation}
where 
$\mathbf{X}_i\in\mathbb{R}^{H\times W\times 3}$ is a pixel-aligned point map for the $i$-th view, 
$\mathbf{K}_i\in\mathbb{R}^{3\times 3}$ is the camera intrinsic matrix, and 
$\mathbf{E}_i\in\mathbb{R}^{3\times 4}$ is the camera extrinsic matrix.
Following VGGT, the predicted point maps are represented in a common coordinate frame defined by the first camera.

\paragraph{Dense facial correspondence.}
To establish dense correspondence between image pixels and the canonical facial UV domain, we use a pretrained Pixel3DMM model~\cite{giebenhain2025pixel3dmm}. 
For each input image $I_i$, Pixel3DMM predicts a UV map:
\begin{equation}
    (\mathbf{U}_i)
    =
    \Phi_{\mathrm{P3D}}(I_i),
\end{equation}
where
\begin{equation}
    \mathbf{U}_i\in[0,1]^{H\times W\times 2}.
\end{equation}
For each image pixel $\mathbf{p}=(x,y)$, 
$\mathbf{U}_i(\mathbf{p})=(u,v)$ indicates its corresponding location in the canonical UV domain.
The UV maps are used to lift VGGT point maps into the canonical UV space.

\subsection{Image-to-UV Geometry Lifting}
\label{sec:image_to_uv_lifting}

The point maps predicted by VGGT provide dense 3D observations in image space, while our target representation is defined in the canonical facial UV domain. 
We therefore lift each per-view point map into UV space using the dense UV correspondence predicted by Pixel3DMM. 
This converts multi-view image observations into a set of partial UV-space point maps that share the same facial parameterization.

\paragraph{Forward splatting to UV space.}
Let $\Omega_i$ denote the set of valid face pixels in the $i$-th image, and let $\Omega_U=\{0,\ldots,S-1\}^2$ denote the discrete UV texel grid. 
For each valid image pixel $\mathbf{p}\in\Omega_i$, the UV map $\mathbf{U}_i$ provides its continuous UV coordinate
\begin{equation}
    \mathbf{U}_i(\mathbf{p})=(u_i(\mathbf{p}),v_i(\mathbf{p}))\in[0,1]^2.
\end{equation}
We forward-splat the corresponding 3D point $\mathbf{X}_i(\mathbf{p})$ to its neighboring UV texels using bilinear weights. 
For a UV texel $\mathbf{q}=(q_x,q_y)\in\Omega_U$, we define
\begin{equation}
    \omega_i(\mathbf{p},\mathbf{q})
    =
    \left[1-\left|q_x-(S-1)u_i(\mathbf{p})\right|\right]_+
    \left[1-\left|q_y-(S-1)v_i(\mathbf{p})\right|\right]_+,
\end{equation}
where $[\cdot]_+=\max(\cdot,0)$.
The accumulated splatting weight is
\begin{equation}
    \mathbf{W}^{uv}_i(\mathbf{q})
    =
    \sum_{\mathbf{p}\in\Omega_i}
    \omega_i(\mathbf{p},\mathbf{q}).
\end{equation}
The partial UV-space point map is then computed by weighted averaging:
\begin{equation}
    \tilde{\mathbf{P}}^{uv}_i(\mathbf{q})
    =
    \frac{
    \sum_{\mathbf{p}\in\Omega_i}
    \omega_i(\mathbf{p},\mathbf{q})
    \mathbf{X}_i(\mathbf{p})
    }{
    \mathbf{W}^{uv}_i(\mathbf{q})+\epsilon
    }.
    \label{eq:image_to_uv_splat}
\end{equation}
We also obtain a UV-space valid mask by thresholding the accumulated weight:
\begin{equation}
    \tilde{\mathbf{A}}^{uv}_i(\mathbf{q})
    =
    \mathbb{1}
    \left[
    \mathbf{W}^{uv}_i(\mathbf{q})>\tau_w
    \right].
\end{equation}

This lifting step produces a partial UV observation 
$(\tilde{\mathbf{P}}^{uv}_i,\tilde{\mathbf{A}}^{uv}_i)$ for each view. 
Regions that are invisible, self-occluded, or outside the facial area remain invalid in the UV domain. 
Thus, the multi-view reconstruction problem is converted into completing and fusing a set of partial UV-space geometric observations.

\paragraph{Canonicalization.}
Although VGGT represents all views in a common coordinate frame, this frame is tied to the input camera system rather than the canonical template, and the resulting point maps may still exhibit residual cross-view misalignment due to imperfect camera and geometry estimation.
To make the UV observations comparable across subjects and sequences, we apply a lightweight per-view canonicalization step. 
Specifically, we sample a sparse set of visible mesh vertices from each partial UV point map using the template UV coordinates, and estimate a similarity transform from these sampled vertices to the canonical template vertices:
\begin{equation}
    \mathcal{T}_i(\mathbf{x})
    =
    s_i \mathbf{R}^{c}_i \mathbf{x}+\mathbf{t}^{c}_i,
\end{equation}
where $s_i$, $\mathbf{R}^{c}_i$, and $\mathbf{t}^{c}_i$ are estimated by Umeyama alignment~\cite{umeyama1991least}. 
We then apply this transform to all valid texels:
\begin{equation}
    \mathbf{P}^{uv}_i(\mathbf{q})
    =
    \mathcal{T}_i
    \left(
    \tilde{\mathbf{P}}^{uv}_i(\mathbf{q})
    \right),
    \quad
    \text{for } \tilde{\mathbf{A}}^{uv}_i(\mathbf{q})=1.
\end{equation}
The valid mask is kept unchanged, except that it is restricted to the valid UV region of the template.

After this step, each input view is represented as a canonical partial UV-space point map:
\begin{equation}
    \left\{
    \left(
    \mathbf{P}^{uv}_i,
    \mathbf{A}^{uv}_i
    \right)
    \right\}_{i=1}^{N}.
\end{equation}
These canonicalized partial observations serve as the input to our UV-space multi-view fusion network.

\subsection{Mask-aware Neural Fusion in UV Space}
\label{sec:uv_fusion_network}

After image-to-UV lifting, each input view is represented as a canonical partial UV-space point map 
$\mathbf{P}^{uv}_i$ with a corresponding valid mask $\mathbf{A}^{uv}_i$. 
Although these partial observations are aligned to the same canonical UV domain, they remain incomplete and noisy due to self-occlusion, inaccurate UV correspondence, and residual errors in the predicted point maps. 
We therefore introduce a neural fusion network to aggregate these partial observations and predict a complete UV-space point map.

\paragraph{Shared per-view encoding.}
A straightforward solution is to directly average the partial point maps in UV space. 
However, raw averaging collapses noisy and incomplete observations before the network can reason about their reliability and local context. 
Instead, we first encode each view independently with a shared encoder. 
For the $i$-th view, we concatenate the canonical UV point map and its valid mask as input:
\begin{equation}
    \mathbf{F}_i
    =
    E_{\theta}
    \left(
    \mathbf{P}^{uv}_i \oplus \mathbf{A}^{uv}_i
    \right),
    \label{eq:shared_uv_encoder}
\end{equation}
where $E_{\theta}$ is a convolutional encoder shared by all views, $\oplus$ denotes channel-wise concatenation, and $\mathbf{F}_i$ is the encoded feature map of the $i$-th partial observation.

The shared encoder has two important roles. 
First, it extracts local geometric context from each partial UV point map before cross-view aggregation. 
Second, because the same encoder is applied to all views, the representation is independent of the input view ordering and can naturally handle a variable number of views.

\paragraph{Mask-aware feature fusion.}
Since each UV point map only covers a subset of the face, invalid texels should not contribute to the fused representation. 
We therefore aggregate the encoded features using the corresponding valid masks. 
Let $\mathbf{A}^{f}_i$ denote the valid mask resized to the spatial resolution of $\mathbf{F}_i$. 
The fused feature map is computed by masked averaging:
\begin{equation}
    \bar{\mathbf{F}}(\mathbf{q})
    =
    \frac{
    \sum_{i=1}^{N}
    \mathbf{A}^{f}_i(\mathbf{q})\,
    \mathbf{F}_i(\mathbf{q})
    }{
    \sum_{i=1}^{N}
    \mathbf{A}^{f}_i(\mathbf{q}) + \epsilon
    },
    \label{eq:masked_feature_fusion}
\end{equation}
where $\mathbf{q}$ indexes a spatial location in the feature map and $\epsilon$ is a small constant for numerical stability. 
This operation only aggregates features from views that provide valid evidence at the corresponding UV location. 
When multiple views observe the same region, their encoded features are averaged; when a region is visible in only one view, the fused feature is determined by that view; and when no view observes a region, the decoder must infer the missing geometry from surrounding context and the learned facial prior.

\paragraph{UV point map completion.}
The fused feature map is then passed to a decoder to predict a complete UV-space point map:
\begin{equation}
    \mathbf{P}^{uv}
    =
    D_{\theta}
    \left(
    \bar{\mathbf{F}}
    \right),
    \label{eq:uv_completion_decoder}
\end{equation}
where $D_{\theta}$ is a convolutional decoder and 
$\mathbf{P}^{uv}\in\mathbb{R}^{S\times S\times 3}$ is the reconstruction point map. 
The decoder completes missing UV regions and regularizes noisy observations into a coherent facial surface. 
The fixed-topology mesh is obtained by sampling $\mathbf{P}^{uv}$ at the template vertex UV coordinates.

\paragraph{Relation to raw UV averaging.}
Our design differs from a simple raw-fusion baseline that first averages the canonical UV point maps and then feeds the averaged point map to a decoder. 
This baseline performs fusion directly in coordinate space and therefore loses view-specific information before neural processing. 
In contrast, our network performs fusion in feature space: each partial observation is first interpreted by the shared encoder, and only then aggregated using valid masks. 
This allows the network to denoise individual observations, exploit local UV context, and learn how to complete unobserved regions more effectively.

\subsection{Training Objectives}
\label{sec:training_objectives}

For each training subject, we bake the ground-truth fixed-topology mesh into the canonical UV domain to obtain a ground-truth UV point map 
$\mathbf{P}^{uv}_{gt}\in\mathbb{R}^{S\times S\times 3}$ 
and its valid mask $\mathbf{A}^{uv}_{gt}$.
Given the predicted reconstruction point map $\mathbf{P}^{uv}$, we use a masked L2 reconstruction loss in UV space:
\begin{equation}
    \mathcal{L}_{pm}
    =
    \frac{
    \sum_{\mathbf{q}}
    \mathbf{A}^{uv}_{gt}(\mathbf{q})
    \left\|
    \mathbf{P}^{uv}(\mathbf{q})
    -
    \mathbf{P}^{uv}_{gt}(\mathbf{q})
    \right\|_2^2
    }{
    \sum_{\mathbf{q}}\mathbf{A}^{uv}_{gt}(\mathbf{q})+\epsilon
    }.
    \label{eq:pm_loss}
\end{equation}

Since the final output is a fixed-topology mesh, we also supervise the vertices sampled from the predicted UV point map. 
Let $\mathbf{V}_{recon}$ and $\mathbf{V}_{gt}$ denote the predicted and ground-truth mesh vertices, respectively. 
The vertex loss is
\begin{equation}
    \mathcal{L}_{v}
    =
    \frac{1}{M}
    \sum_{j=1}^{M}
    \left\|
    \mathbf{v}_{recon,j}
    -
    \mathbf{v}_{gt,j}
    \right\|_2^2.
    \label{eq:vertex_loss}
\end{equation}

To reduce local surface artifacts, we further match the Laplacian coordinates of the predicted and ground-truth meshes. 
Let $\mathbf{L}$ be the uniform mesh Laplacian constructed from the fixed topology $\mathbf{F}$. 
The Laplacian matching loss is
\begin{equation}
    \mathcal{L}_{lap}
    =
    \frac{1}{M}
    \sum_{j=1}^{M}
    \left\|
    \left(\mathbf{L}\mathbf{V}_{recon}\right)_j
    -
    \left(\mathbf{L}\mathbf{V}_{gt}\right)_j
    \right\|_1.
    \label{eq:laplacian_loss}
\end{equation}
This term encourages the reconstructed mesh to preserve the local differential structure of the ground-truth surface, which helps suppress spurious bumpy artifacts.

The full objective for the neural fusion network is
\begin{equation}
    \mathcal{L}
    =
    \lambda_{pm}\mathcal{L}_{pm}
    +
    \lambda_{v}\mathcal{L}_{v}
    +
    \lambda_{lap}\mathcal{L}_{lap}.
    \label{eq:base_total_loss}
\end{equation}

\section{EXPERIMENTS}
\begin{table*}[htbp]
\centering
\caption{Quantitative comparison on public datasets (all metrics in mm, $\downarrow$ indicates lower is better).}
\label{tab:quantitative}
\resizebox{\textwidth}{!}{
\begin{tabular}{l c c c c c c c c c c c c}
\toprule
\multirow{2}{*}{Method} & 
\multicolumn{1}{c}{H3DS} & 
\multicolumn{3}{c}{NeRSemble} & 
\multicolumn{3}{c}{EmoTalk3D} & 
\multicolumn{3}{c}{RenderMe-360} & 
\multirow{2}{*}{Time} &
\multirow{2}{*}{\# Views} \\
\cmidrule(lr){2-2} 
\cmidrule(lr){3-5}
\cmidrule(lr){6-8}
\cmidrule(lr){9-11}
& Mean $\downarrow$ 
& Mean $\downarrow$ & Median $\downarrow$ & Std $\downarrow$ 
& Mean $\downarrow$ & Median $\downarrow$ & Std $\downarrow$ 
& Mean $\downarrow$ & Median $\downarrow$ & Std $\downarrow$ 
& & \\
\midrule
DECA & 1.99 & 1.58 & 1.55 & 1.18 & 1.56 & 1.51 & 1.23 & 1.71 & 1.65 & 1.24 & $< 1s$ & 1 \\
3DDFA-V3 & 1.81 & 1.52 & 1.46 & 1.08 & 1.51 & 1.46 & 1.19 & 1.72 & 1.65 & 1.27 & $< 1s$ & 1 \\
Pixel3DMM & 1.69 & 1.42 & 1.38 & 1.01 & 1.31 & 1.28 & 0.99 & 1.48 & 1.38 & 1.06 & $\sim 60s$ & 1 \\
DFNRMVS & 1.78 & 1.62 & 1.58 & 1.09 & 1.72 & 1.67 & 1.30 & 1.77 & 1.68 & 1.25 & $\sim 30s$ & 16 \\
HRN & 1.46 & 1.49 & 1.43 & 1.10 & 1.64 & 1.57 & 1.28 & 1.65 & 1.55 & 1.30 & $\sim 70s$ & 16 \\
VGGTFace & 1.18 & 0.98 & 0.98 & \textbf{0.65} & 1.02 & 0.99 & 0.76 & 1.08 & 0.98 & 0.74 & $< 10s$ & 16 \\
\midrule
Ours & \textbf{1.12} & \textbf{0.91} & \textbf{0.91} & 0.69 & \textbf{0.89} & \textbf{0.86} & \textbf{0.74} & \textbf{0.93} & \textbf{0.84} & \textbf{0.72} & $< 3s$ & 16 \\
w/o Neural Fusion & 1.22 & 1.01 & 0.99 & 0.76 & 1.01 & 0.98 & 0.82 & 1.03 & 0.91 & 0.83 & $< 3s$ & 16 \\
\bottomrule
\end{tabular}
}
\vspace{-1.5em}
\end{table*}

\subsection{Experimental Setup}
\label{sec:experimental_setup}

\paragraph{Implementation details.}
All input images are resized to $512\times512$, and the UV-space point maps are also represented at a resolution of $512\times512$. 
We train our networks on Ava-256~\cite{martinez2024ava256}. 
We use the official ground-truth meshes and register them to the FLAME topology, which provides the fixed mesh connectivity and canonical UV parameterization used by our method. 
During training, we randomly sample $4$--$16$ input views for each subject. 
The neural fusion network is optimized using AdamW. 
It is trained for 200 epochs on two RTX 4090 GPUs for about two days.

\paragraph{Datasets.}
We evaluate our method on four datasets: H3DS~\cite{ramon2021h3d,caselles2025implicit}, NeRSemble~\cite{kirschstein2023nersemble}, EmoTalk3D~\cite{he2024emotalk3d}, and RenderMe-360~\cite{pan2024renderme360}. 
We additionally capture several in-the-wild multi-view sequences for evaluation. 
Unless otherwise stated, all methods are evaluated with 16 input views. 
For EmoTalk3D, we use 11 views since only 11 camera views are available. 
We also report 4-view quantitative and qualitative results in the supplementary material.

\paragraph{Baselines.}
We compare with both monocular and multi-view face reconstruction methods. 
The monocular baselines include DECA~\cite{DECA:Siggraph2021}, 3DDFA-V3~\cite{wang20243d}, and Pixel3DMM~\cite{giebenhain2025pixel3dmm}. 
For these methods, we run inference on all input views and report the best reconstruction. 
The multi-view baselines include DFNRMVS~\cite{bai2020deep}, HRN~\cite{lei2023hierarchical}, and VGGTFace~\cite{ming2026vggtface}. 
We also include an ablation variant of our method that removes neural feature fusion: it directly computes a masked average of the canonical UV-space point maps and feeds the averaged point map to the same decoder.

\paragraph{Metrics.}
We use Chamfer distance as the main quantitative metric and report all errors in millimeters. 
For H3DS and NeRSemble, we follow the evaluation protocol used by prior work and quote the numbers reported in SIRA++~\cite{caselles2025implicit} and VGGTFace. 
For EmoTalk3D and RenderMe-360, we reconstruct ground-truth geometry with Metashape\footnote{https://www.agisoftmetashape.com} using all available camera views provided by the datasets. 
Following the evaluation protocol of SIRA++, we align each predicted mesh to the corresponding ground-truth mesh using a similarity transformation followed by rigid ICP before computing the Chamfer distance.
We also compare runtime across methods. 
For our method, the reported runtime includes VGGT inference, Pixel3DMM inference, UV-space point-map construction, neural fusion network inference, and final mesh extraction. 
All runtime measurements are conducted on a single RTX 4090 GPU.

\subsection{Quantitative Comparison}
\label{sec:quantitative_results}

Table~\ref{tab:quantitative} reports the quantitative comparison on four evaluation datasets. 
Our method achieves the lowest Chamfer distance on almost all reported metrics, including the mean error on H3DS and the mean/median errors on NeRSemble, EmoTalk3D, and RenderMe-360. 
Compared with monocular methods, our method benefits from multi-view geometric observations and produces more accurate fixed-topology reconstructions. 
Compared with multi-view baselines, our canonical UV-space representation more effectively aggregates partial observations from different views.

Table~\ref{tab:quantitative} also reports the runtime of different methods. 
Our full pipeline takes less than 3 seconds on a single RTX 4090.
While some monocular methods are faster, they use only a single image and have noticeably higher reconstruction errors. 
Among multi-view methods, our method achieves both higher accuracy and faster inference, without requiring per-subject optimization.

\subsection{Qualitative Comparison}
\label{sec:qualitative_results}

\begin{figure}[p]
    \centering
    \includegraphics[
        width=\textwidth,
        height=0.975\textheight,
        keepaspectratio
    ]{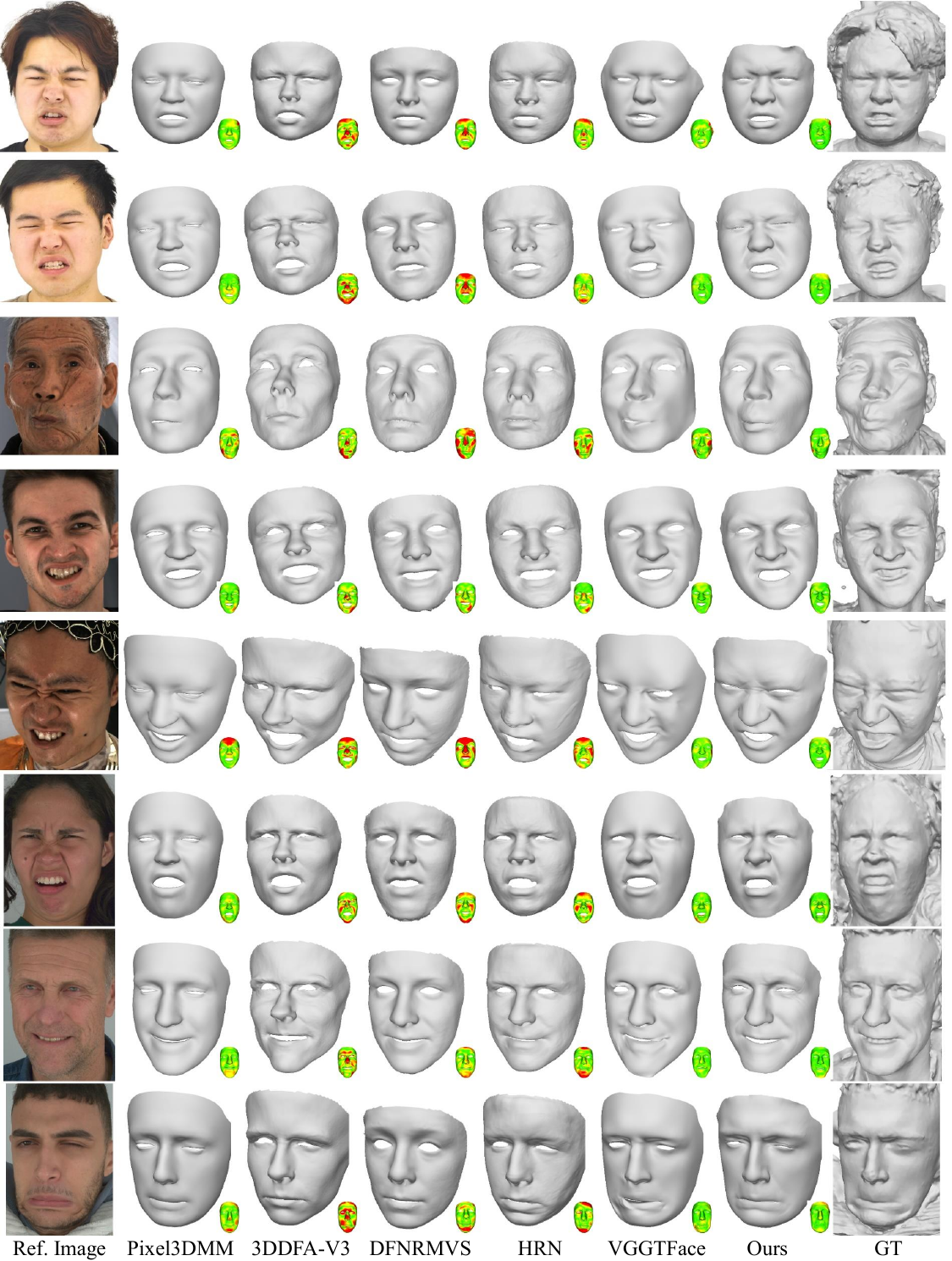}
    \caption{
    \textbf{Qualitative comparisons on public datasets.}
    Our method reconstructs cleaner fixed-topology meshes and better preserves challenging expression geometry around the eyes, mouth, and cheeks.
    Error maps are shown beside each reconstruction.
    }
    \label{fig:qualitative_cmp}
\end{figure}

\begin{figure}[p]
    \centering
    \includegraphics[
        width=0.89\textwidth,
        height=\textheight,
        keepaspectratio
    ]{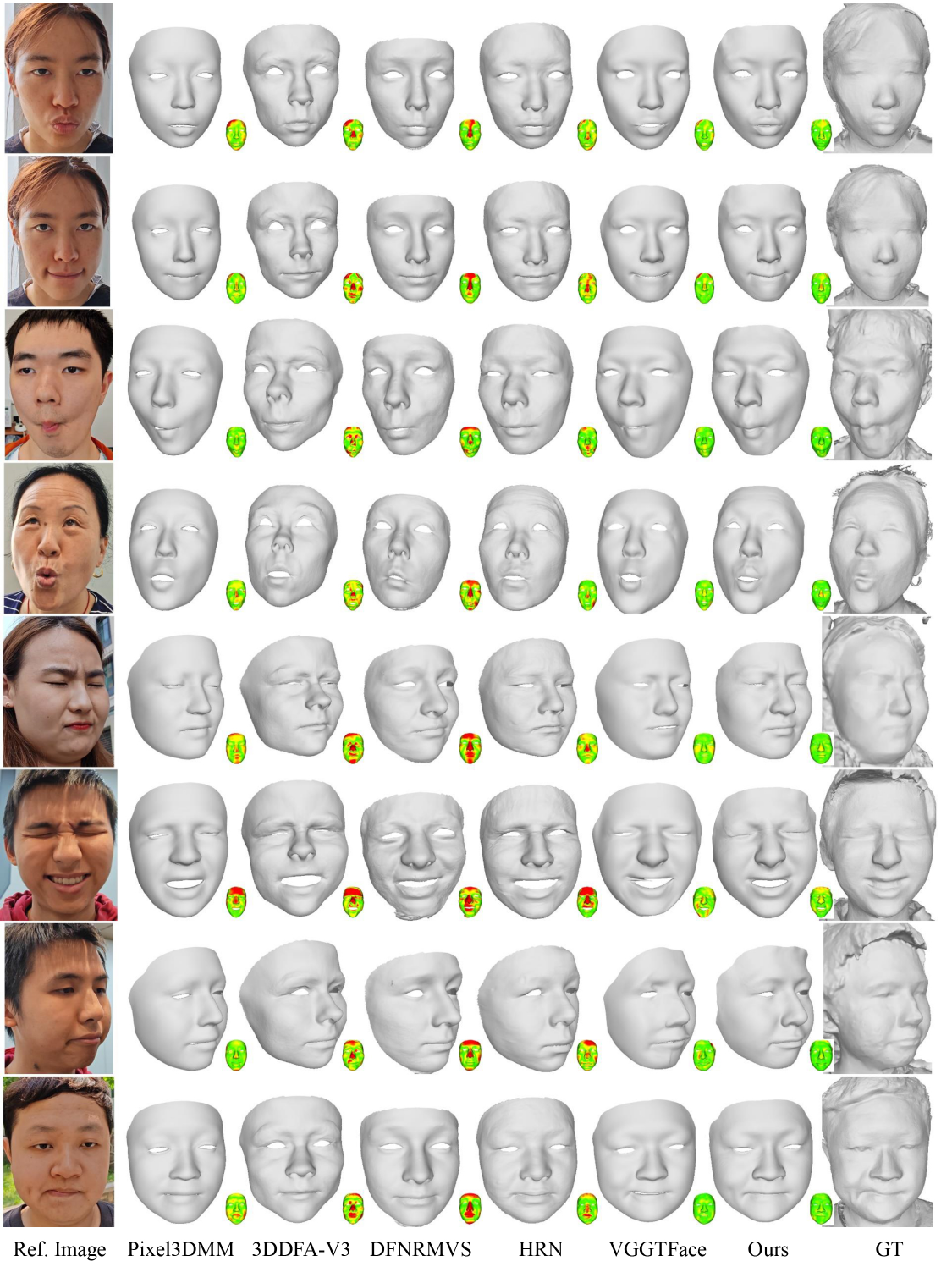}
    \caption{
    \textbf{Qualitative comparisons on in-the-wild captures.}
    Our method generalizes to casual multi-view images and produces stable, faithful face meshes across diverse identities, viewpoints, lighting conditions, and expressions.
    Error maps are shown beside each reconstruction.
    }
    \label{fig:qualitative_cmp_inthewild}
\end{figure}

Figure~\ref{fig:qualitative_cmp} and Figure~\ref{fig:qualitative_cmp_inthewild} show qualitative comparisons on public datasets and in-the-wild captures. 
We highly urge the reader to check our \emph{suppl. video} for more comparison results. 
Monocular methods, such as Pixel3DMM and 3DDFA-V3, can recover plausible coarse face shapes, but their reconstructions are often biased toward generic facial priors. 
As a result, they tend to miss subject-specific traits and struggle with challenging expressions, such as tightly closed eyes, puffed cheeks, asymmetric smiles, and large mouth deformations. 
Multi-view methods improve geometric consistency by using more observations, but DFNRMVS and HRN can still produce noisy surfaces, distorted local structures, or inaccurate expression geometry.

Compared with these baselines, our method reconstructs cleaner fixed-topology meshes with more faithful global shape and local facial deformation. 
In particular, our results better preserve expression-related geometry around the eyes, mouth, cheeks, and nasolabial regions, while avoiding the surface noise and local artifacts produced by some multi-view baselines. 
VGGTFace benefits from strong VGGT geometry and produces competitive results, but its heuristic fusion can over-smooth local facial structures or fail to fully resolve inconsistent observations. 
The in-the-wild results further demonstrate the generalization ability of our approach. 
Although trained on Ava-256, our method remains robust under casual capture conditions with different lighting, viewpoints, identities, and expressions. 

\subsection{User Study}
\begin{figure}[t]
    \centering
    \includegraphics[width=0.45\textwidth]{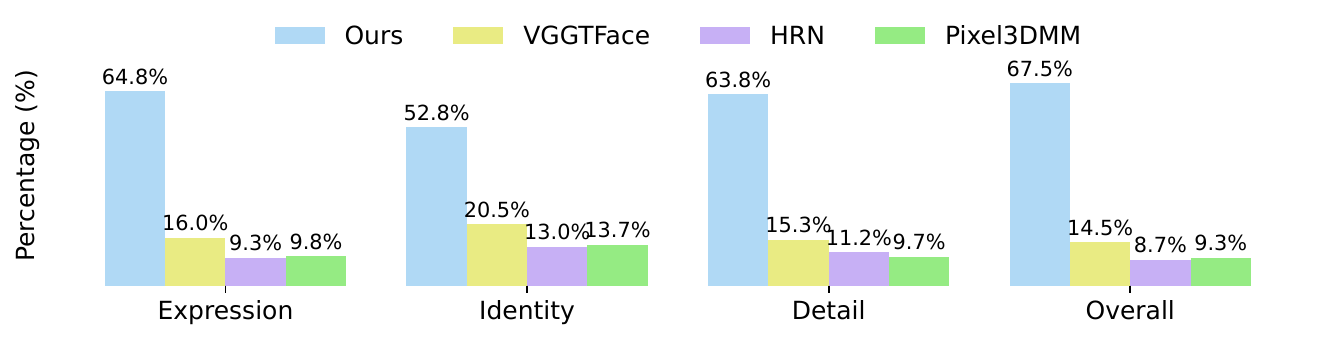}
    \caption{
        User study results. Participants compared the reconstruction results of Pixel3DMM, HRN, VGGTFace, and our method without seeing the method names. Our method receives the highest preference across expression matching, identity preservation, detail preservation, and overall quality.
}
    \label{Fig:user_study}
    \vspace{-1em}
\end{figure}

To further evaluate the perceptual quality of the reconstructed meshes, we conducted a user study comparing our method with Pixel3DMM, HRN, and VGGTFace. Each participant was shown the input multi-view images and the reconstructed meshes produced by the four methods. The method names were hidden during the study. For each example, the four results were randomly assigned to options A, B, C, and D to avoid ordering bias.

For each test case, participants were asked to answer the following four questions:
(1) Which reconstruction best matches the facial expression in the input images?
(2) Which reconstruction best matches the identity-related facial features, such as face shape and facial components?
(3) Which reconstruction best preserves the facial details in the input images?
(4) Overall, which reconstruction result is the best? 
We provide a screenshot of the web interface used in our user study in the supplementary material.

We invited 25 participants, and each participant evaluated 24 groups of reconstruction results. Therefore, each criterion received 600 votes in total. The final voting percentages are shown in Fig.~\ref{Fig:user_study}. Our method is preferred by the majority of participants across all four criteria.
These results indicate that the advantages of our method are not only reflected in numerical errors, but are also clearly perceived by human observers.

\subsection{Ablation Study}
\label{sec:ablation}

\begin{figure}[tbp]
    \centering
    \includegraphics[width=0.4\textwidth]{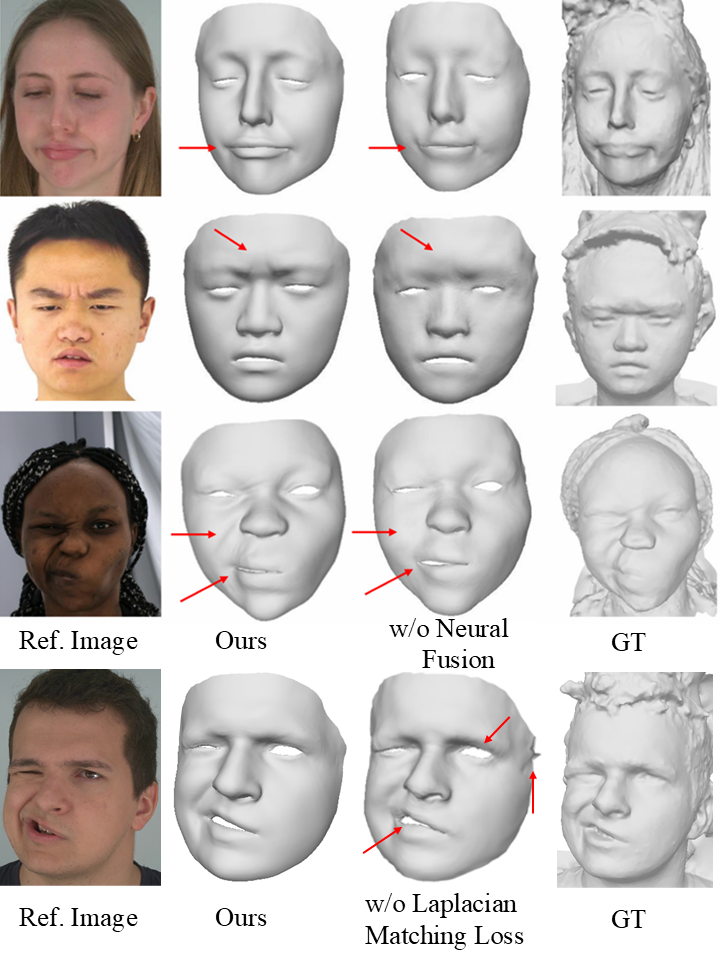}
    \caption{
        Qualitative evaluation of the key design choices in our method.
        Without neural fusion, directly averaging canonical UV-space point maps loses view-specific evidence before decoding, leading to less accurate expressions and degraded local facial geometry.
        Without the Laplacian matching loss, the reconstructed mesh captures the overall shape but exhibits less regular surfaces and visible bumpy artifacts.
    }
    \label{Fig:ablation}
    \vspace{-2em}
\end{figure}

We analyze two important design choices in our method: neural feature fusion and Laplacian matching. 
As shown in Table~\ref{tab:quantitative}, removing neural fusion and directly averaging canonical UV-space point maps leads to worse quantitative performance on all four datasets. 
This suggests that raw coordinate-space averaging is insufficient for handling noisy and partially inconsistent observations.

Figure~\ref{Fig:ablation} further shows the qualitative effect of this design. 
Without neural fusion, the decoder receives an already-collapsed UV point map, where view-specific evidence and local reliability cues have been lost. 
As a result, the reconstructed mesh can have less accurate expressions and weaker local geometry.

We also ablate the Laplacian matching loss. 
Without this term, the reconstructed mesh still captures the overall face shape, but the surface becomes less regular and exhibits visible bumpy artifacts.

\subsection{Limitations and Discussions}
\label{sec:limitations}

\begin{figure}[t]
    \centering
    \includegraphics[width=0.45\textwidth]{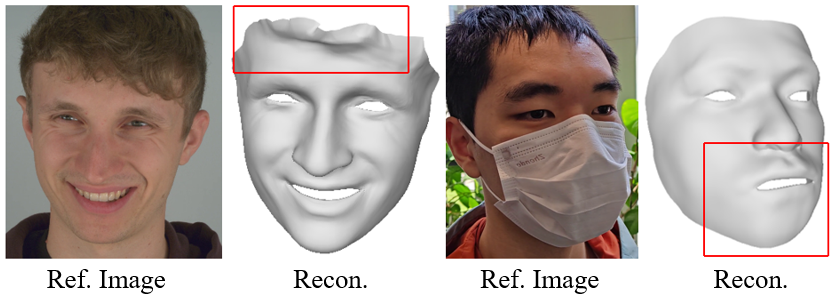}
    \caption{
        Limitations of our method. 
        Our method can absorb bangs into the facial mesh when hair geometry is present in the predicted point maps or registered training meshes.
        It also struggles with severe occlusions such as masks, where UV correspondences become unreliable and the occluder is reconstructed as part of the visible geometry.
    }
    \label{Fig:limit}
    \vspace{-2em}
\end{figure}

Figure~\ref{Fig:limit} shows typical failure cases of our method. 
First, our reconstruction may include bangs or hair-like geometry near the forehead. 
This happens because both the VGGT point maps and the registered training meshes may preserve such geometry, causing the fixed-topology face surface to absorb the bangs into the reconstructed mesh.
Second, our method struggles with strong occlusions such as face masks. 
In these cases, Pixel3DMM may provide inaccurate UV correspondences around the occluded region, while VGGT reconstructs the occluder as part of the visible geometry. 
Since our representation does not explicitly separate occluders from the face surface, the final reconstruction can fail under severe occlusion. Incorporating explicit face--hair and face--occluder decomposition could prevent non-face geometry from being absorbed into the fixed-topology facial surface. 
In addition, our current system reconstructs each capture as a static subject. 
Extending the proposed canonical UV-space fusion framework to dynamic sequences, with temporal consistency and motion-aware fusion, would be an interesting direction.

\section{CONCLUSIONS}

We presented UVFaceFusion, a feed-forward framework for fixed-topology face reconstruction from in-the-wild multi-view images. 
By lifting VGGT point maps into a canonical facial UV space with Pixel3DMM correspondences, our method turns topologically consistent reconstruction into UV point-map completion and replaces heuristic optimization with mask-aware neural fusion. 
Experiments demonstrate our method obtains state-of-the-art results on various benchmarks and strong generalization ability on in-the-wild data captured by everyday users.

\bibliographystyle{ACM-Reference-Format}
\bibliography{sample-bibliography}

\clearpage
\onecolumn

\clearpage

\clearpage

\appendix

\twocolumn[
\begin{center}
    {\LARGE\bfseries
    UVFaceFusion: Fast Multi-view Topologically Consistent Face Reconstruction
    in the Wild via UV-space Neural Fusion\par}
    \vspace{0.5em}
    {\large\itshape Supplementary Material\par}
    \vspace{1.5em}
\end{center}
]

\section{Quantitative Evaluation with Four Input Views}

\begin{table*}[t]
\centering
\caption{Quantitative comparison using four input views. Lower values are better. Our method consistently achieves the lowest reconstruction errors across all datasets while maintaining the fastest inference speed.}
\label{tab:comparison_4view}
\resizebox{\textwidth}{!}{
\begin{tabular}{l c c c c c c c c c c c c}
\toprule
\multirow{2}{*}{Method} & 
\multicolumn{1}{c}{H3DS} & 
\multicolumn{3}{c}{NeRSemble} & 
\multicolumn{3}{c}{EmoTalk3D} & 
\multicolumn{3}{c}{RenderMe-360} & 
\multirow{2}{*}{Time} &
\multirow{2}{*}{\# Views} \\
\cmidrule(lr){2-2} 
\cmidrule(lr){3-5}
\cmidrule(lr){6-8}
\cmidrule(lr){9-11}
& Mean $\downarrow$ 
& Mean $\downarrow$ & Median $\downarrow$ & Std $\downarrow$ 
& Mean $\downarrow$ & Median $\downarrow$ & Std $\downarrow$ 
& Mean $\downarrow$ & Median $\downarrow$ & Std $\downarrow$ 
& & \\
\midrule
DFNRMVS & 1.78 & 1.63 & 1.58 & 1.10 & 1.74 & 1.67 & 1.31 & 1.80 & 1.71 & 1.26 & $\sim 5s$ & 4 \\
HRN & 1.56 & 1.57 & 1.56 & 1.16 & 1.73 & 1.61 & 1.34 & 1.83 & 1.74 & 1.36 & $\sim 15s$ & 4 \\
VGGTFace & 1.37 & 1.09 & 1.05 & 0.80 & 1.21 & 1.11 & 0.90 & 1.28 & 1.15 & 0.91 & $\sim 3s$ & 4 \\
\midrule
Ours & \textbf{1.31} & \textbf{1.06} & \textbf{1.02} & \textbf{0.74} & \textbf{1.02} & \textbf{1.01} & \textbf{0.73} & \textbf{1.12} & \textbf{1.04} & \textbf{0.84} & $\sim 1s$ & 4 \\
\bottomrule
\end{tabular}
}
\end{table*}

In the main paper, we report quantitative comparisons under the 16-view input setting. To further evaluate the performance of our method under a more sparse-view scenario, we conduct an additional quantitative experiment using only four input views. The evaluation protocol, metrics, and test datasets are the same as those used in the main paper. The results are reported in Table~\ref{tab:comparison_4view}. Lower values indicate better reconstruction accuracy.

As shown in Table~\ref{tab:comparison_4view}, our method consistently achieves the best performance across all evaluated datasets under the 4-view setting. 
The results demonstrate that our method remains effective even when the number of input views is significantly reduced. This robustness mainly benefits from our UV-space fusion strategy, which aggregates partial observations from different views using valid masks, and from the learned shape prior that completes unobserved regions in the canonical UV domain. Moreover, our method runs at approximately 1 second per subject, which is faster than the compared methods while producing more accurate reconstructions.

\section{Qualitative Evaluation with Four Input Views}

\begin{figure*}[p]
    \centering
    \includegraphics[
        width=0.9\textwidth,
        height=\textheight,
        keepaspectratio
    ]{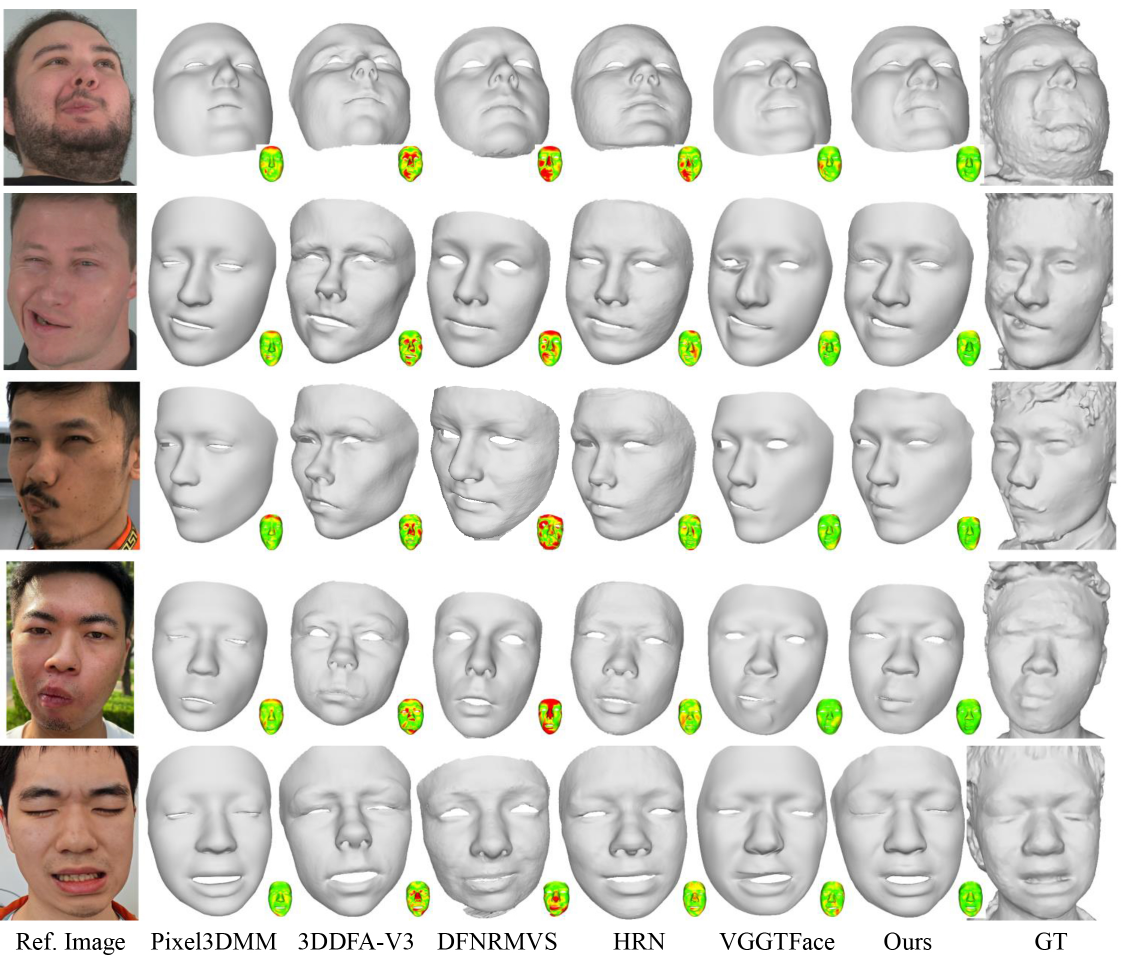}
    \caption{
    Qualitative comparison under the 4-view input setting. 
    For compactness, we show one reference image from the input views in the first column, while all methods use the same four input views. 
    From left to right, we show the results of Pixel3DMM, 3DDFA-V3, DFNRMVS, HRN, VGGTFace, our method, and the ground truth. 
    Compared with the baselines, our method reconstructs more accurate facial expressions and identity-related geometry, while producing more complete and stable fixed-topology meshes under sparse-view inputs.
    }
    \label{fig:supp_4view_qual}
\end{figure*}

We further provide qualitative comparisons under the 4-view input setting in Fig.~\ref{fig:supp_4view_qual}. The examples include both public benchmark datasets, namely H3DS, NeRSemble, EmoTalk3D, and RenderMe-360, and in-the-wild captures collected by ourselves. All methods are evaluated using the same sparse-view inputs.

The 4-view setting is more challenging than the 16-view setting because large portions of the face may be weakly observed or unobserved. As a result, directly fusing sparse observations or relying on per-view predictions can easily lead to incomplete geometry, inaccurate facial expressions, or over-smoothed surfaces. In contrast, our method reconstructs a complete fixed-topology face mesh by completing the partial UV-space point maps in a canonical domain. The qualitative results show that our method better preserves the input expression and identity-related facial geometry, while producing more stable and detailed surfaces. These results are consistent with the quantitative improvements reported in Table~\ref{tab:comparison_4view}.

\section{More Details on Our Captured Data}

\begin{figure}[htbp]
    \centering
    \includegraphics[width=0.475\textwidth]{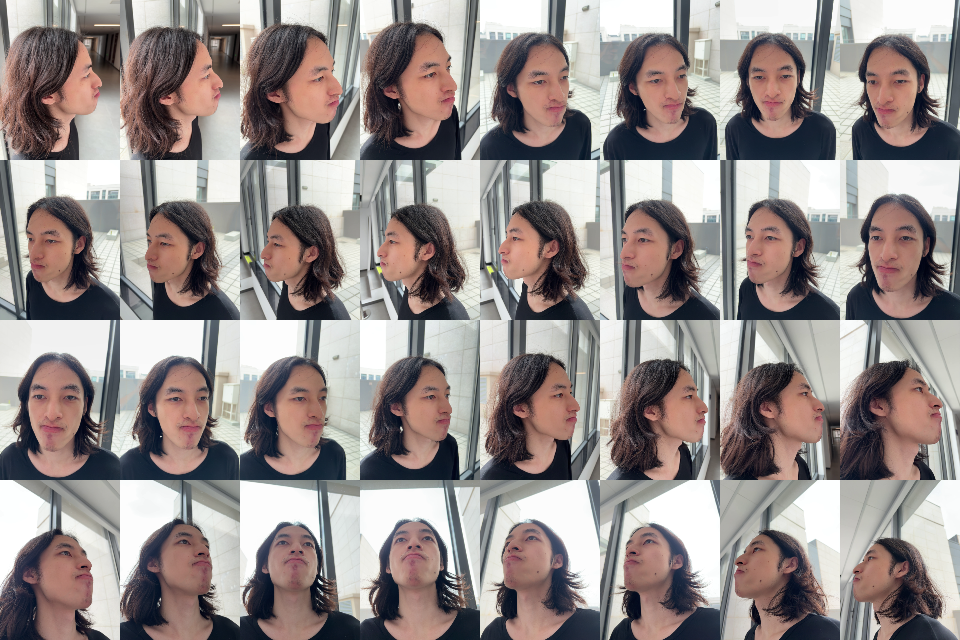}
    \caption{
        Montage of densely sampled frames from one of our self-captured in-the-wild videos.
    }
    \label{Fig:grid}
\end{figure}

We provide additional details about our self-captured in-the-wild data. 
The videos are recorded with a hand-held smartphone, including both iPhone and Android devices, at a resolution of $1920 \times 1080$. 
For each subject, we capture a short video by moving the phone around the subject's head at a distance of approximately $0.8$--$1.5$ meters. 
For the 16-view setting, we uniformly sample 16 frames from each video and use them as the multi-view input to our system. 
For visualization, Fig.~\ref{Fig:grid} shows a denser sampling of frames from one captured video, illustrating the camera trajectory and the viewpoint coverage during capture.

\section{Monocular Reconstruction}

\begin{figure}[htbp]
    \centering
    \includegraphics[width=0.35\textwidth]{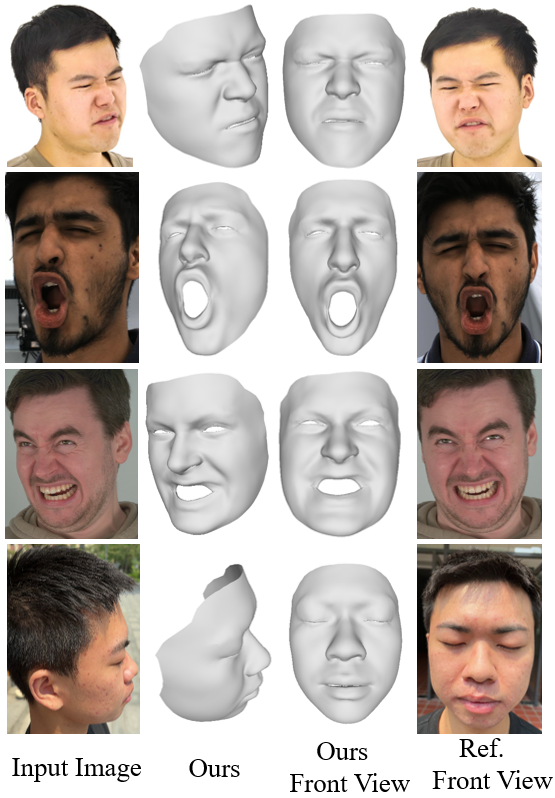}
    \caption{
        Monocular reconstruction results. Although our method is mainly designed for multi-view inputs, it can be extended to the single-view setting by using one partial UV-space point map and completing the missing regions in the canonical UV domain. These results demonstrate the flexibility of our formulation. Note, our method only uses the input image in the first column; the front-view image in the last column is shown only as a reference and is not used for reconstruction.
    }
    \label{Fig:mono_recon}
\end{figure}

Although our method is designed for multi-view face reconstruction, it can also be naturally extended to the monocular setting. VGGT can predict a reasonable point map from a single input image. We then lift the predicted point map into the UV space using the UV map estimated by Pixel3DMM, obtaining a partial UV-space point map and its corresponding valid mask. In this case, our reconstruction network takes only one partial UV-space point map as input, and completes the missing regions using the learned facial shape prior in the canonical UV domain.

We show several monocular reconstruction results in Fig.~\ref{Fig:mono_recon}. In the monocular setting, our method takes only the input image as input, while the front-view image is provided solely for visual reference and is not used by our method. Since monocular reconstruction is not the main focus of this work, we do not include quantitative comparisons or baseline comparisons in this setting. Nevertheless, the results demonstrate that our UV-space formulation is flexible and can handle extremely sparse observations. Even with only a single view, our method can reconstruct a complete fixed-topology face mesh and produce plausible geometry for the unobserved regions.

\section{User Study Screenshot}
\begin{figure*}[t]
    \centering
    \includegraphics[width=0.95\textwidth]{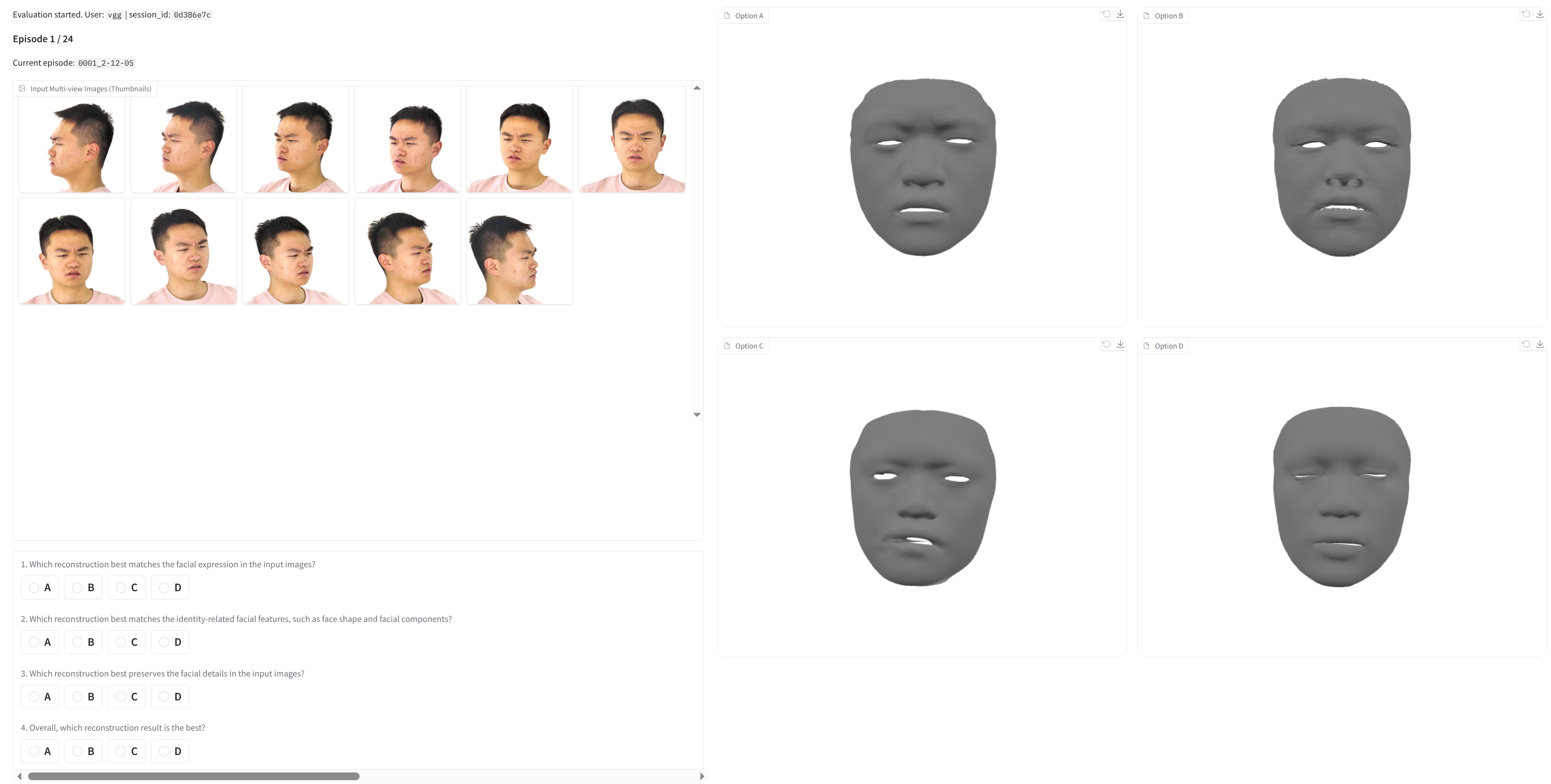}
    \caption{
        Screenshot of the web interface used in our user study. 
        Participants are shown the input multi-view images and four anonymized reconstruction results. 
        Each mesh viewer is interactive, allowing participants to zoom in and rotate the mesh for closer inspection. 
        Participants then select the best result for expression, identity, detail, and overall quality.
    }
    \label{fig:user_study_interface}
\end{figure*}

We provide a screenshot of the web interface used in our user study, as shown in Fig.~\ref{fig:user_study_interface}. 
The left panel shows the input multi-view images and the four evaluation questions, while the right panel shows the reconstruction results from four anonymized methods, denoted as Options A--D. 
The four mesh viewers are interactive: participants can zoom in and drag the meshes to inspect the reconstructed geometry from different viewpoints before making their choices. 
The method names are not displayed, and the assignment of methods to Options A--D is randomized for each test case.

\end{document}